\documentclass[runningheads]{llncs}
\usepackage{xcolor}    
\usepackage{eccv}

\usepackage{eccvabbrv}

\usepackage{graphicx}
\usepackage{booktabs}
\usepackage{amsmath,amssymb,amsfonts}
\usepackage[table]{xcolor}
\usepackage{colortbl}
\usepackage{multirow}
\usepackage{array}
\usepackage{rotating}
\usepackage{float}
\usepackage[hypcap=false]{caption}
\usepackage{enumitem}
\usepackage[accsupp]{axessibility}  

\usepackage{hyperref}

\usepackage{orcidlink}

\definecolor{best}{RGB}{198,239,206}
\definecolor{second}{RGB}{255,235,156}
\definecolor{third}{RGB}{255,199,206}
\definecolor{clrGranulation}{RGB}{255,0,0}
\definecolor{clrSlough}{RGB}{255,255,0}
\definecolor{clrNecrosis}{RGB}{128,0,128}
\definecolor{clrOutside}{RGB}{0,0,0}
\definecolor{clrSkinRemnant}{RGB}{0,76,255}
\definecolor{clrTendon}{RGB}{0,229,255}
\definecolor{clrBackground}{RGB}{0,0,0}
\definecolor{clrSkin}{RGB}{0,0,255}
\definecolor{clrWound}{RGB}{255,0,0}
\newcommand{\colorswatch}[1]{\colorbox{#1}{\phantom{\rule{1.2ex}{1.2ex}}}}

\begin{document}

\title{A Deep Learning Benchmark for Wound Tissue Segmentation}
\titlerunning{A Deep Learning Benchmark for Wound Tissue Segmentation}
\author{Can Rollas\inst{1} \and
Mehmet Kemal Güllü\inst{2,3} \and
İbrahim Onur Alıcı\inst{2,3} \and
Bayram Çolak\inst{2,3} \and
Yalın Baştanlar\inst{1}}
\institute{Department of Computer Engineering, Izmir Institute of Technology, Izmir, Türkiye \and
Amatis Inc., Izmir, Türkiye \and
Izmir Bakırçay University, Izmir, Türkiye\\
\email{canrollas@iyte.edu.tr} and 
\email{yalinbastanlar@iyte.edu.tr}}
\authorrunning{C. Rollas et al.}
\maketitle

\begin{abstract}
Pixel-level classification of wound tissue types --granulation, slough/fibrin, and necrosis-- is a clinically important yet understudied problem, and progress has been hampered by the absence of large, publicly available annotated datasets. We introduce a clinical dataset of 568 pixel-annotated RGB images collected from Çiğli Training and Research Hospital (Bakircay University), which we publicly release and, to our knowledge, is the largest publicly available dataset for intra-wound tissue segmentation. Building on this resource, we present a systematic benchmark of 37 configurations spanning 9 segmentation architectures, 12 encoder backbones from six families, and 3 pretrained weight strategies, evaluated under a strictly controlled protocol to isolate architectural effects. Our key finding is that the encoder backbone is the dominant performance driver. Vision Transformer backbones (MiT series) outperform all CNN families regardless of segmentation architecture choice, while pretrained weights play a secondary role. We also provide a benchmark for wound region segmentation (background/skin/wound) across 29 configurations with varying architectures, backbones and pretrained weights. This second task is performed on a publicly available dataset consisting 2760 images. We further offer deployment recommendations covering accuracy--efficiency trade-offs for clinical settings. The dataset is available at \url{https://collab.amatis.tr/en}.

\end{abstract}

\section{Introduction}
\label{sec:intro}

\begin{figure}[tb]
  \centering
  \includegraphics[width=\linewidth]{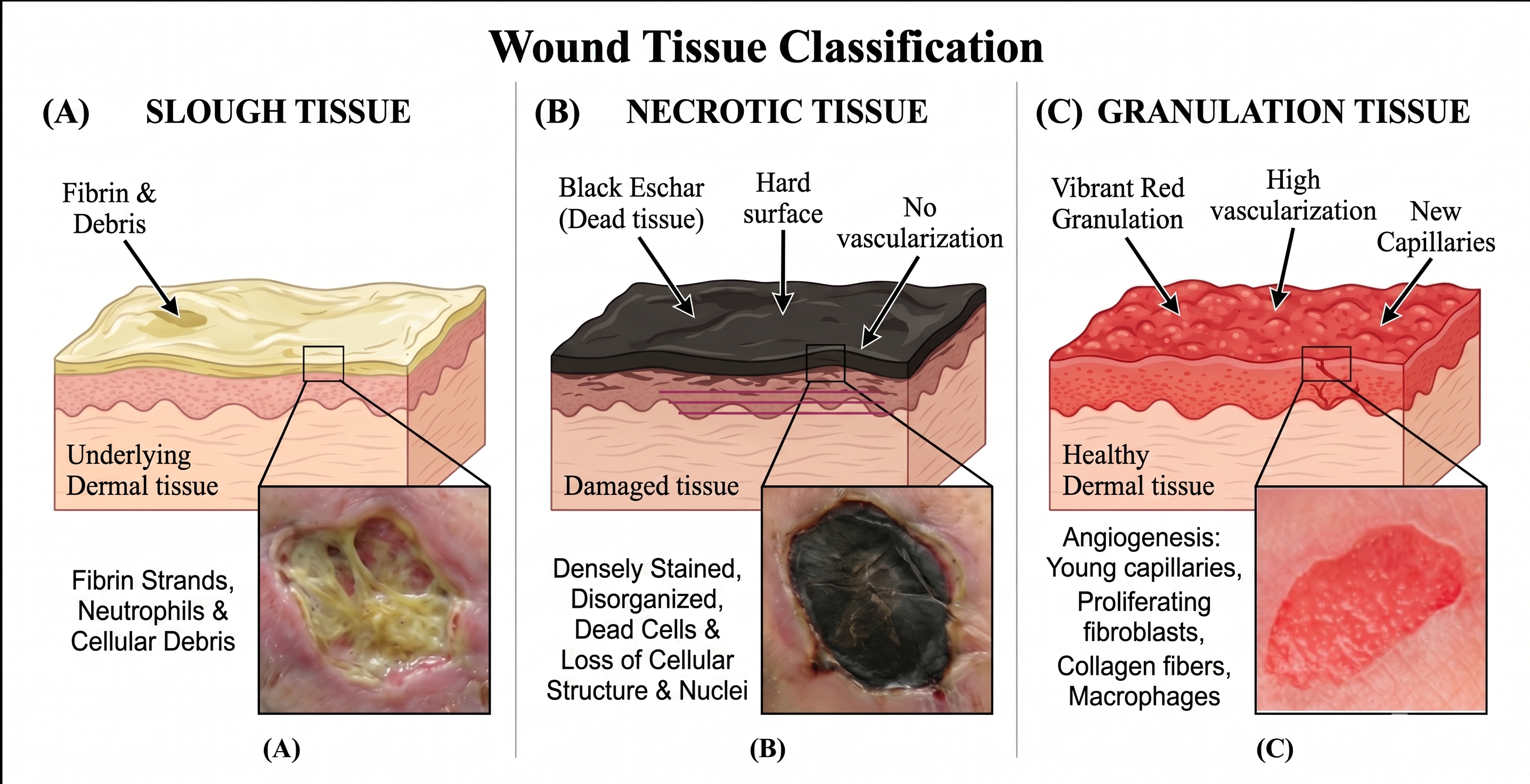}
  \caption{Schematic illustrations and key features of (A)~slough, (B)~necrosis,
           and (C)~granulation tissues in wound bed.}
  \label{fig:tissue_types}
\end{figure}

Chronic wounds (including pressure ulcers, diabetic foot
ulcers, and venous leg ulcers) impose a substantial clinical and economic burden, with tens of millions of patients affected worldwide and treatment costs
representing a major burden on healthcare systems \cite{sen2019burden}. The composition of tissue within the wound bed is a principal determinant of healing trajectory and treatment appropriateness:
\emph{granulation tissue}, characterised by highly vascularised connective tissue with a distinctively red, granular appearance, indicates active neo-vascularisation and a favourable healing milieu;
\emph{slough} (also termed fibrin or yellow slough), a mixture of dead leucocytes and fibrin, impedes epithelialisation and typically mandates debridement; \emph{necrosis} (eschar), consisting of desiccated devitalised tissue, is associated with deep tissue injury and severely compromises healing unless
promptly removed. Accurate, reproducible measurement of the relative proportions of these tissue types enables clinicians to stratify wound severity, select debridement
modalities, monitor healing response, and predict time to closure. \cref{fig:tissue_types} illustrates the histological and visual characteristics of each tissue class.

Conventional wound assessment relies on clinician visual inspection and manual planimetry, a process that is time-consuming and subject to substantial inter- and intra-rater variability. As deep learning has prevailed in many detection and segmentation tasks in the medical domain \cite{Kahveci2026, Kahveci2024, Park2023, Yigit2022}, wound assessment has also taken advantage of automated pixel-level segmentation using convolutional neural networks (CNNs) \cite{goyal2020wound}. More recently, Vision Transformer architectures offer a path towards standardized and scalable wound tissue segmentation.

Despite the clear clinical motivation, a systematic benchmarking study for this specific task is lacking. Existing work either evaluates a single architecture in isolation \cite{goyal2020dfuseg}, 
employs small and non-public datasets \cite{wu2022autotissue},
or addresses only whole-wound region detection without extending to intra-wound tissue segmentation \cite{kuo2024fusegnet}.
Consequently, practitioners deploying wound assessment systems lack principled guidance on which architectural choices most strongly influence performance. To address this gap, we design a comprehensive benchmark under a controlled experimental protocol, explicitly separating the contributions of segmentation architecture, encoder backbone, and pretrained weights.

Our work spans two complementary segmentation tasks.
First, \textit{wound region segmentation}: a three-class segmentation (background, surrounding skin, and wound region) that delineates the wound from the broader body surface. Second, \textit{intra-wound tissue segmentation}: fine-grained pixel-level classification of the wound interior into granulation, slough, and necrosis. Our novel tissue benchmark dataset, a clinical corpus of 568 annotated RGB images, excludes the surrounding skin since wound region segmentation is addressed at the previous stage.

The specific contributions of this work are as follows:
\begin{enumerate}[leftmargin=*,noitemsep]
  \item Employing the novel tissue segmentation dataset, the first systematic benchmarking for intra-wound tissue segmentation, evaluating 37 configurations across segmentation architecture, encoder backbone, and pretrained weight axes under a strictly controlled protocol.
  \item Empirical demonstration that Mix Transformer (MiT) encoders substantially outperform all tested CNN encoder families, independent of the segmentation architecture choice.
  \item A wound region segmentation benchmark across 29 configurations on a publicly available collection of 2{,}760 wound images, which we extend with manual three-class (background/skin/wound) annotations and publicly release, constituting the first stage of the pipeline through which the wound region is delineated from surrounding skin and background prior to tissue-level analysis.
  \item Evidence-based deployment recommendations which cover accuracy--efficiency trade-offs to guide model selection in clinical settings.
\end{enumerate}

The remainder of this paper is structured as follows.
\cref{sec:related} reviews the wound segmentation literature and the general-purpose architectures and backbone families evaluated in this work.
\cref{sec:dataset} describes the wound datasets used in this study. \cref{sec:method} details the training protocol and the evaluation metrics. \cref{sec:results} presents experimental results. Finally, 
\cref{sec:conclusions} concludes the paper.

\section{Related Work}
\label{sec:related}

Automated wound assessment using deep learning can be decomposed into two complementary tasks: \textit{wound region segmentation}, which delineates the wound perimeter from surrounding healthy skin, and \textit{intra-wound tissue segmentation}, which classifies pixels within the wound bed into distinct
biological categories.

\subsection{Wound Region Segmentation}

The encoder-decoder paradigm introduced by U-Net \cite{ronneberger2015unet} remains the most widely adopted framework for wound region segmentation. Applied to a limited corpus of diabetic foot ulcer (DFU) images,
a standard U-Net demonstrated strong feature extraction capabilities, achieving 97.25\% Dice Score and 94.86\% IoU~\cite{goyal2020dfuseg} (cf. Table \ref{tab:litreview-wb}).

On a heterogeneous real-world dataset of 362 wound images captured by nurses
under varying illumination conditions,
U-Net attained 92.7\% Dice Score and 86.8\% IoU, though the authors noted that
unmasked full-resolution inputs occasionally caused the network to confuse healthy reddish tissue in the background with wound
tissue~\cite{mcnichols2026cnn}.
These results confirm U-Net's strength in controlled datasets while highlighting its vulnerability to distribution shift.

To improve robustness against background clutter, attention mechanisms have
been integrated into the U-Net framework.
A Dual Attention U-Net built on a VGG16 backbone combines spatial and
channel-wise attention gates that suppress irrelevant background activations
and focus decoding on wound  pixels~\cite{zhu2025dualattn}.
Trained on DFU images and fine-tuned on acute and chronic wound data,
this model achieved 94.1\% Dice Score and 89.3\% IoU, demonstrating strong
cross-dataset generalisation.

Feature Pyramid Networks (FPN)~\cite{lin2017fpn} address the multi-scale
nature of wound lesions by fusing features at multiple resolutions.
The WSNet framework combines local patch-level and global whole-image
predictions through a convolutional fusion layer, with FPN achieving the best
balance at 84.7\% Dice Score \cite{goyal2020wound}.
DeepLabV3+ augmented with a Squeeze-and-Excite (SE) channel recalibration
module on a ResNet-50 backbone dynamically weights feature channel
contributions, yielding 92.3\% Dice and 92.4\% IoU while also providing
reliable wound perimeter and major/minor axis length
estimates~\cite{iyer2023woundparam}.

FUSegNet~\cite{kuo2024fusegnet} is a single-stage encoder--decoder network
that pairs a pretrained EfficientNet-B7 encoder with a decoder embedding a
modified parallel spatial and channel squeeze-and-excitation (P-scSE) module at the middle of each decoder stage, combining additive and max-out scSE to better recalibrate features for wound delineation. On a publicly available chronic wound dataset, it attains a data-based Dice score of 92.70\%.

The best scores obtained for this task were reported in \cite{goyal2020dfuseg}, however only 650 images were used (a part of DFU dataset).We extend this benchmark to a larger and more heterogeneous collection of 2{,}760 wound images (cf. Sec.~\ref{sec:woundregion}).

\begin{table}[H]
\caption{Wound Region Segmentation: Summary of reported results in the literature. 
${\dagger}$~IoU derived from the reported Dice Score, and vice-versa.} 
\label{tab:litreview-wb}
\renewcommand{\arraystretch}{1.1}
\begin{tabular*}{\linewidth}{@{\extracolsep{\fill}}lrrrl@{}}
\toprule
\textbf{Architecture} & \textbf{IoU (\%)} & \textbf{Dice (\%)} & \textbf{Year} & \textbf{Ref.} \\
\midrule
Dual Attention U-Net (VGG16)         & 89.3             & 94.1  & 2025 & \cite{zhu2025dualattn} \\
WSNet (FPN backbone)                 & 73.5$^{\dagger}$ & 84.7  & 2023 & \cite{goyal2020wound} \\
DeepLabV3+ \& SE (ResNet-50)          & 92.4             & 92.3  & 2023 & \cite{iyer2023woundparam} \\
FUSegNet (EfficientNet-B7)           & 86.4$^{\dagger}$ & 92.7  & 2024 & \cite{kuo2024fusegnet} \\
U-Net (real-world, heterogeneous)    & 86.8             & 92.7  & 2026 & \cite{mcnichols2026cnn} \\
U-Net (DFU, limited dataset)         & 94.9             & 97.3  & 2020 & \cite{goyal2020dfuseg} \\
U-Net (Swift wound dataset)          & 86.4               & 92.7$^{\dagger}$ & 2022 & \cite{wu2022autotissue} \\
\bottomrule
\end{tabular*}
\end{table}

\subsection{Intra-Wound Tissue Segmentation}

Classifying pixels within the wound bed into tissue types such as granulation, slough, fibrin, necrosis, maceration, and epithelial tissue is substantially
harder than binary wound delineation.
Tissue boundaries are defined by subtle colour and texture gradients rather than sharp edges, and significant inter-rater disagreement exists among expert
annotators~\cite{wu2022autotissue}.
Class imbalance is severe: necrosis and maceration are often rare, small, or co-located with other tissue types. Due to these challenges, the number of previous intra-wound segmentation studies is low.

A comprehensive evaluation ~\cite{kabir2025fpnvgg} constructed a dataset of 147 wound images spanning six tissue classes annotated in three labelling formats (full image, patch, and superpixel), then comparing model configurations including U-Net and FPN variants.
FPN combined with a VGG16 feature extractor achieved the highest weighted average Dice Score of 82.25\%, outperforming all tested segmentation architecture and encoder backbone combinations.
This work presents a systematic backbone-level benchmark, however the dataset is very limited and not publicly available.


\begin{table}[t]
\centering
\caption{Intra-Wound Tissue Segmentation: Summary of reported results in the literature.
${\dagger}$~IoU derived from the reported Dice Score, and vice-versa.} 
\label{tab:litreview-wt}
\begin{tabular}{lrrrr}
\toprule
\textbf{Architecture} & \textbf{IoU (\%)} & \hspace{2mm} \textbf{Dice (\%)}  & \hspace{2mm} \textbf{Year} & \hspace{2mm} \textbf{Ref.} \\
\midrule
FPN (VGG16)                          & 69.9$^{\dagger}$ & 82.25 & 2025 & \cite{kabir2025fpnvgg} \\
U-Net (EfficientNet-B0) & 71.92      & $83.65^{\dagger}$   & 2022 & \cite{wu2022autotissue} \\
\midrule
\end{tabular}
\end{table}

To bring tissue quantification to bedside devices, the AutoTissue
system~\cite{wu2022autotissue} combined a wound region segmentation module
with a lightweight tissue
segmentation module based on EfficientNet-B0.

Dhar~\emph{et al.}~\cite{dhar2024woundtissue} conducted a pilot study on
wound tissue segmentation in diabetic foot ulcer images using deep learning,
further highlighting the challenges of fine-grained tissue classification in
clinical settings. However, they concentrated on semi-supervised learning and used a dataset of only 110 images labelled by the domain experts. Whereas our benchmarking is on supervised learning with a larger dataset completely labeled by experts.


\section{Dataset}
\label{sec:dataset}

We use two datasets: a publicly available dataset used for the wound region segmentation stage and a clinical intra-wound tissue segmentation dataset introduced for the first time in this work.

\subsection{Wound Region Segmentation Dataset}
\label{sec:woundregion}

This dataset consists of 2{,}760 wound images ($512\times512$ pixels) obtained from a publicly available Kaggle release~\cite{leoscode_kaggle} that aggregates three public wound image sources: the Foot Ulcer Segmentation Challenge (FUSeg)~\cite{wang2020data,wang2024fuseg} (1{,}210 images), the WSNet dataset~\cite{goyal2020wound} (1{,}176 images), and the Medetec Wound Database~\cite{thomas2020medetec} (374 images).
Source membership was determined from the filename prefixes used in the released collection.
As the WSNet and Medetec portions contain chronic wounds of various aetiologies, the collection is broader than diabetic foot ulcers alone.
The aggregated release provides binary wound masks only.
We extend it to a three-class pixel-level annotation (background/skin/wound) by manually delineating the surrounding body surface, and publicly release this three-class version~\cite{rollas_kaggle}.
The images are split into 2{,}208 training and 552 test images.
A distinguishing characteristic of this annotation scheme is the explicit labelling of surrounding skin as a dedicated class.
In most wound datasets, only the wound region is annotated; here, skin is separately delineated because reliable wound region segmentation requires the model
to understand the body surface context surrounding the wound, not merely to detect the wound in isolation.
This dataset is used only for the wound region segmentation stage of the clinical pipeline, where a 29-configuration benchmark determined the optimal wound region segmentation model. 

\subsection{Intra-Wound Tissue Dataset}
\label{sec:woundtissue}

The tissue dataset used in this benchmark consists of 568 RGBA images ($512\times512$ pixels) collected from clinical wound assessment sessions. \footnote{The dataset is publicly available at
\url{https://collab.amatis.tr/en}}
Each image uses its alpha channel as a wound-region mask: pixels with
alpha~$=$~255 lie inside the wound and carry a tissue-class label encoded as a
RGB colour in the accompanying annotation mask, while pixels with alpha~$=$~0
lie outside the wound and are assigned a void label (value~255) excluded from
both loss computation and metric evaluation.
Three biologically meaningful tissue classes are defined as detailed in \cref{tab:classes}.
Although additional pixel categories were present during annotation, pixels labelled as skin remnant
(blue) and tendon (cyan) appear in too few images to support reliable
learning and are therefore treated as void and excluded together with outside-wound pixels (black).

\begin{figure}[H]
\centering
\includegraphics[width=\linewidth]{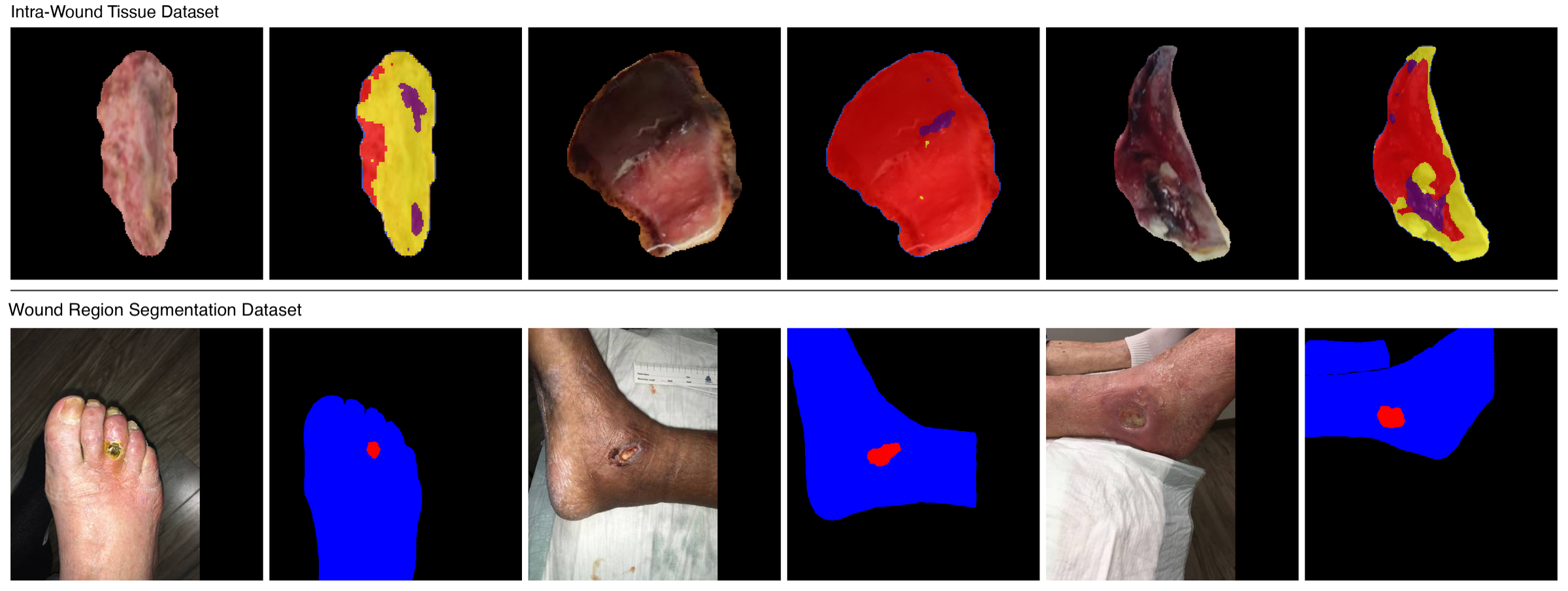}
\caption{Representative samples from both datasets.
\textit{Top row}: intra-wound tissue dataset samples with annotation masks
(red~=~granulation, yellow~=~slough/fibrin, purple~=~necrosis).
\textit{Bottom row}: wound region segmentation dataset samples
(black~=~background, blue~=~skin, red~=~wound region).
Best viewed in colour.} 

\label{fig:dataset_samples}
\end{figure}

\noindent
\begin{minipage}[t]{1.0\linewidth}
\centering
\captionof{table}{Intra-Wound Tissue Class Definitions}
\label{tab:classes}
\begin{tabular}{clc}
\toprule
\textbf{Label} & \textbf{Tissue Type} & \textbf{Colour} \\
\midrule
0   & Granulation      & \colorswatch{clrGranulation}  \\
1   & Slough/Fibrin    & \colorswatch{clrSlough}       \\
2   & Necrosis         & \colorswatch{clrNecrosis}     \\
\midrule
255 & Outside wound    & \colorswatch{clrOutside}\ (ignored)   \\
255 & Skin remnant     & \colorswatch{clrSkinRemnant}\ (ignored) \\
255 & Tendon           & \colorswatch{clrTendon}\ (ignored)    \\
\bottomrule
\end{tabular}
\end{minipage}
\hfill

\vspace{1em}

\section{Methodology}
\label{sec:method}

Our methodology comprises two components: a systematic benchmarking protocol that evaluates segmentation architectures and encoder backbones on intra-wound tissue data, and an end-to-end two-stage clinical pipeline that integrates the best-performing model into a complete wound analysis workflow.

In Stage~1, a wound region segmentation model crops the wound region of interest from
the input image; in Stage~2, the tissue segmentation model classifies pixels
within the cropped region into granulation, slough/fibrin, and necrosis.
The following subsections detail the architectures, backbones, pretrained
weights, and training protocol used across both stages.

\subsection{Segmentation Architectures}

We evaluate nine segmentation architectures using Segmentation Models PyTorch~(SMP~v0.5.0)~\cite{yakubovskiy2019smp}, which provides standardised implementations sharing a common encoder backbone API, enabling fair
architectural comparison.
The nine architectures evaluated are U-Net~\cite{ronneberger2015unet},
UNet++~\cite{zhou2018unetplusplus}, FPN~\cite{lin2017fpn},
DeepLabV3+~\cite{chen2018deeplabv3plus}, MA-Net~\cite{fan2020manet},
LinkNet~\cite{chaurasia2017linknet}, PAN~\cite{li2018pan},
PSPNet~\cite{zhao2017pspnet}, and DeepLabV3~\cite{chen2017deeplabv3}.

\subsection{Encoder Backbones}

Twelve encoder backbones spanning six families are evaluated.
For each architecture, only encoder--decoder pairings confirmed to be compatible
by the SMP library are included, yielding 37 valid configurations in total.
Incompatible combinations (\textit{e.g.} UNet++ \& MiT, DeepLabV3 \& Xception)
are excluded.
The encoders are:

\begin{itemize}[noitemsep]
  \item \text{ResNet family}: ResNet-18, ResNet-34, ResNet-50
        \cite{he2016resnet}
  \item \text{EfficientNet family}: EfficientNet-B0, B3, B4
        \cite{tan2019efficientnet}
  \item \text{Lightweight CNN}: MobileNetV2 \cite{sandler2018mobilenetv2}
  \item \text{Dense CNN}: DenseNet-121 \cite{huang2017densenet}
  \item \text{Grouped convolution}: ResNeXt-50(32$\times$4d)
        \cite{xie2017resnext}
  \item \text{Transformer}: MiT-B1, MiT-B2, MiT-B3
        \cite{xie2021segformer}
\end{itemize}

\subsection{Pretrained Weight Strategies}

Three pretrained weight transfer options are compared, where semi-supervised pretrained weights are only available for ResNet-50 encoders.
\begin{itemize}[noitemsep]
  \item \text{ImageNet}: Weights pretrained on ImageNet-1K.
  \item \text{SSL}: Semi-supervised learning on 1~billion weakly-labelled
        images \cite{yalniz2019ssl}.
  \item \text{SWSL}: Combined supervised and semi-supervised learning
        \cite{yalniz2019ssl}.
\end{itemize}

\subsection{Data Partitioning and Pre-processing}

The 568 images are partitioned using 3-fold cross-validation with a fixed random seed of~42.
Images are first shuffled with this seed, then divided into three consecutive chunks of
$\lfloor 568/3 \rfloor = 189$ images each (the final fold receives the one remaining image, giving 190).
Each fold uses its chunk as the test set; 10\% of the remaining images form the validation set
(37~images, used for checkpoint selection based on highest mean~IoU); the rest form the training set
(342~images for folds~1 and~2; 341~for fold~3).
All reported metrics are means over the three test folds.

\textbf{Image resizing.}
The intra-wound tissue dataset is natively stored at $512\times512$ pixels and
is fed to the models at that resolution without downsampling.
The wound region segmentation dataset is also stored at $512\times512$
but is downsampled to $256\times256$ at load time for comparability with prior work reported on the FUSeg data.
Bilinear interpolation is used for RGB images; nearest-neighbour interpolation
is used for annotation masks to preserve discrete label values without
introducing spurious interpolated colours at class boundaries.


\textbf{Normalisation.}
Pixel values are scaled to $[0,1]$ and then standardised using ImageNet
channel statistics:
$\mu = [0.485,\;0.456,\;0.406]$,
$\sigma = [0.229,\;0.224,\;0.225]$.

\textbf{Data Augmentation.}
Spatial augmentations are applied online during training and identically to
each image--mask pair to preserve label consistency.
Random horizontal flip and random vertical flip are each applied independently
with probability~$p=0.5$.
A random rotation from $\{90^\circ,\;180^\circ,\;270^\circ\}$ is applied with
probability~$p=0.7$.
No colour or photometric augmentation is employed, as tissue colour is
diagnostically meaningful in both annotation schemes.

\subsection{Training Protocol}

All 37 configurations share an identical training protocol to ensure that observed performance differences are attributable solely to architectural
choices:

\begin{itemize}[noitemsep]
  \item \text{Optimiser}: AdamW~\cite{loshchilov2019adamw} with
        $\text{lr} = 10^{-4}$, weight decay $= 10^{-5}$.
  \item \text{Learning rate schedule}: Cosine annealing over 100
        epochs~\cite{loshchilov2016cosine} with
        $\eta_{\min} = \text{lr} \times 0.01$.
  \item \text{Loss function}:
        $\mathcal{L} = 0.5\,\mathcal{L}_{\text{CE}} +
                       0.5\,\mathcal{L}_{\text{Dice}}$,
        where $\mathcal{L}_{\text{CE}}$ is cross-entropy and
        $\mathcal{L}_{\text{Dice}}$ is the  Dice
        loss .
        Both losses exclude void-labelled pixels (label value~255).
  \item \text{Epochs}: 100, with checkpoint selection based on best mean~IoU on the per-fold validation set (see \cref{sec:dataset}).
\end{itemize}

\section{Experimental Results}
\label{sec:results}

Experiment results are given in the same order. First, Stage 1, wound region segmentation and then Stage 2, intra-wound tissue segmentation. 

We report per-class Intersection over Union (IoU), per-class Dice coefficient,
mean~IoU~(mIoU), and mean~Dice~(mDice) computed over the three tissue classes.
\begin{align}
\text{IoU}_{c}   &= \frac{|P_c \cap G_c|}{|P_c \cup G_c|}\\[4pt]
\text{Dice}_{c}  &= \frac{2|P_c \cap G_c|}{|P_c| + |G_c|}
\end{align}

\noindent where $P_c$ and $G_c$ denote the predicted and ground-truth binary masks for class $c \in \{0,1,2\}$. 

\cref{tab:wound_loc_benchmark} reports all 29 Stage~1 wound region segmentation
configurations benchmarked on the wound region segmentation dataset.
\cref{tab:full} reports all 37
Stage-2 configurations benchmarked on our dataset.

\begin{table}[H]
\centering
\footnotesize
\caption{Wound region segmentation benchmark on the wound region segmentation dataset. mIoU and mDice are reported as mean{\scriptsize$\pm$std}, with std estimated by non-parametric bootstrap ($n=1000$) on the test set. IoU\textsubscript{BG}, IoU\textsubscript{S}, and IoU\textsubscript{W} denote background, skin, and wound IoU. Best result highlighted. Params: model parameters in millions; T(ms): single-image GPU inference latency. DLV3+ stands for DeepLabV3+.}
\label{tab:wound_loc_benchmark}
\setlength{\tabcolsep}{4pt}
\resizebox{\linewidth}{!}{%
\begin{tabular}{lllcccrrrrr}
\toprule
\textbf{Archite.} & \textbf{Backbone} & \textbf{Weights}
  & \textbf{Params} & \textbf{mIoU} & \textbf{mDice}
  & \textbf{IoU\textsubscript{BG}} & \textbf{IoU\textsubscript{S}} & \textbf{IoU\textsubscript{W}} & \textbf{T(ms)} \\
\midrule
\rowcolor{best}
U-Net         & MiT-B2           & ImageNet &  27.48 & \textbf{0.9001}{\scriptsize$\pm$.0050} & \textbf{0.9439}{\scriptsize$\pm$.0028} & 0.9911 & 0.9604 & 0.7488 & 5.12 \\
U-Net         & ResNet-50        & SSL      &  32.52 & 0.8999{\scriptsize$\pm$.0051} & 0.9439{\scriptsize$\pm$.0028} & 0.9902 & 0.9579 & 0.7515 & 3.32 \\
U-Net++       & DenseNet-121     & ImageNet &  30.07 & 0.8998{\scriptsize$\pm$.0045} & 0.9432{\scriptsize$\pm$.0025} & 0.9910 & 0.9599 & 0.7460 & 11.90 \\
U-Net         & MiT-B1           & ImageNet &  16.43 & 0.8984{\scriptsize$\pm$.0048} & 0.9429{\scriptsize$\pm$.0027} & 0.9905 & 0.9587 & 0.7460 & 3.16 \\
U-Net++       & ResNet-50        & ImageNet &  48.99 & 0.8984{\scriptsize$\pm$.0055} & 0.9428{\scriptsize$\pm$.0031} & 0.9912 & 0.9603 & 0.7438 & 10.78 \\
MA-Net        & MiT-B2           & ImageNet &  34.97 & 0.8970{\scriptsize$\pm$.0051} & 0.9420{\scriptsize$\pm$.0028} & 0.9904 & 0.9584 & 0.7422 & 5.81 \\
FPN           & MiT-B2           & ImageNet &  26.08 & 0.8969{\scriptsize$\pm$.0049} & 0.9421{\scriptsize$\pm$.0027} & 0.9893 & 0.9556 & 0.7459 & 5.19 \\
U-Net         & DenseNet-121     & ImageNet &  13.61 & 0.8957{\scriptsize$\pm$.0049} & 0.9412{\scriptsize$\pm$.0027} & 0.9902 & 0.9573 & 0.7396 & 5.28 \\
DLV3+         & ResNet-50        & SWSL     &  26.68 & 0.8953{\scriptsize$\pm$.0048} & 0.9410{\scriptsize$\pm$.0027} & 0.9894 & 0.9553 & 0.7411 & 2.61 \\
DLV3+         & MiT-B3           & ImageNet &  45.23 & 0.8952{\scriptsize$\pm$.0051} & 0.9410{\scriptsize$\pm$.0028} & 0.9896 & 0.9561 & 0.7399 & 8.46 \\
U-Net++       & EfficientNet-B3  & ImageNet &  13.63 & 0.8946{\scriptsize$\pm$.0048} & 0.9405{\scriptsize$\pm$.0027} & 0.9896 & 0.9561 & 0.7399 & 8.46 \\
MA-Net        & ResNet-50        & ImageNet & 147.44 & 0.8946{\scriptsize$\pm$.0051} & 0.9405{\scriptsize$\pm$.0028} & 0.9902 & 0.9570 & 0.7365 & 5.34 \\
U-Net         & ResNet-50        & SWSL     &  32.52 & 0.8945{\scriptsize$\pm$.0050} & 0.9403{\scriptsize$\pm$.0028} & 0.9899 & 0.9564 & 0.7376 & 9.57 \\
U-Net++       & ResNet-34        & ImageNet &  26.08 & 0.8943{\scriptsize$\pm$.0054} & 0.9402{\scriptsize$\pm$.0030} & 0.9906 & 0.9583 & 0.7347 & 3.31 \\
U-Net         & EfficientNet-B4  & ImageNet &  20.23 & 0.8942{\scriptsize$\pm$.0049} & 0.9402{\scriptsize$\pm$.0027} & 0.9898 & 0.9562 & 0.7344 & 4.62 \\
U-Net         & ResNet-50        & ImageNet &  32.52 & 0.8931{\scriptsize$\pm$.0053} & 0.9394{\scriptsize$\pm$.0030} & 0.9904 & 0.9574 & 0.7315 & 5.61 \\
U-Net         & EfficientNet-B3  & ImageNet &  13.16 & 0.8918{\scriptsize$\pm$.0049} & 0.9387{\scriptsize$\pm$.0027} & 0.9895 & 0.9547 & 0.7311 & 3.36 \\
MA-Net        & ResNet-34        & ImageNet &  31.78 & 0.8900{\scriptsize$\pm$.0052} & 0.9375{\scriptsize$\pm$.0029} & 0.9897 & 0.9549 & 0.7253 & 4.69 \\
U-Net         & ResNet-34        & ImageNet &  24.44 & 0.8897{\scriptsize$\pm$.0053} & 0.9374{\scriptsize$\pm$.0030} & 0.9896 & 0.9547 & 0.7247 & 3.17 \\
MA-Net        & EfficientNet-B3  & ImageNet &  17.17 & 0.8876{\scriptsize$\pm$.0046} & 0.9361{\scriptsize$\pm$.0026} & 0.9885 & 0.9515 & 0.7228 & 2.62 \\
FPN           & EfficientNet-B4  & ImageNet &  19.35 & 0.8874{\scriptsize$\pm$.0044} & 0.9362{\scriptsize$\pm$.0025} & 0.9876 & 0.9496 & 0.7251 & 5.51 \\
DLV3+         & ResNet-50        & ImageNet &  26.68 & 0.8872{\scriptsize$\pm$.0050} & 0.9359{\scriptsize$\pm$.0028} & 0.9884 & 0.9515 & 0.7216 & 5.80 \\
U-Net         & ResNet-18        & ImageNet &  14.33 & 0.8857{\scriptsize$\pm$.0051} & 0.9348{\scriptsize$\pm$.0029} & 0.9891 & 0.9525 & 0.7154 & 2.61 \\
FPN           & ResNeXt-50       & ImageNet &  25.59 & 0.8850{\scriptsize$\pm$.0045} & 0.9345{\scriptsize$\pm$.0025} & 0.9883 & 0.9503 & 0.7164 & 1.87 \\
FPN           & ResNet-50        & ImageNet &  26.12 & 0.8846{\scriptsize$\pm$.0048} & 0.9343{\scriptsize$\pm$.0027} & 0.9879 & 0.9496 & 0.7164 & 2.56 \\
U-Net         & EfficientNet-B0  & ImageNet &   6.25 & 0.8815{\scriptsize$\pm$.0050} & 0.9321{\scriptsize$\pm$.0028} & 0.9884 & 0.9483 & 0.7060 & 2.65 \\
U-Net         & MobileNet-V2     & ImageNet &   6.63 & 0.8809{\scriptsize$\pm$.0057} & 0.9317{\scriptsize$\pm$.0032} & 0.9881 & 0.9489 & 0.7056 & 2.95 \\
FPN           & ResNet-34        & ImageNet &  23.16 & 0.8802{\scriptsize$\pm$.0051} & 0.9314{\scriptsize$\pm$.0029} & 0.9873 & 0.9425 & 0.7057 & 2.02 \\
DLV3+         & EfficientNet-B3  & ImageNet &  11.68 & 0.8745{\scriptsize$\pm$.0047} & 0.9276{\scriptsize$\pm$.0027} & 0.9869 & 0.9453 & 0.6912 & 4.62 \\\bottomrule
\end{tabular}}
\end{table}

Stage 1 dataset (wound region segmentation 2{,}760 images) is pre-divided into
2{,}208 training and 552 test images. Thus, instead of $k$-fold cross-validation
we performed non-parametric \textit{resampling with replacement} (bootstrap).
In each of the $1{,}000$ replicates, 552 samples (i.e., the original test set
size) are drawn with replacement from the test set and the metrics are recomputed; we report (in \cref{tab:wound_loc_benchmark}) the mIoU and mDice standard deviations across those $1{,}000$ replicates. The resampling is seeded
(seed $=42$) so that all 29 configurations are evaluated on identical replicate
index sets, making the reported deviations directly comparable.

For Stage 1, best ranked model is U-Net with MiT-B2 (ImageNet weights, mIoU~$=$~0.9001). For Stage~2 (intra-wound tissue segmentation), best-performing  model is MA-Net (Multi-scale Attention Network) with
MiT-B2 encoder (mIoU $=$ 0.7263, mDice~$=$~0.8407).
For deployment under tighter constraints, U-Net\& MiT-B1 offers the best
accuracy--efficiency balance (mIoU~$=$~0.7203 at lower parameter count than MiT-B2
configurations), while U-Net\&EfficientNet-B0 or U-Net\&MobileNetV2 can be
preferred when model size is the primary constraint.

\begin{table}[H]
\caption{Intra-Wound Tissue Segmentation: Full Benchmark Results Ranked by Mean~IoU (3-fold cross-validation mean; mIoU and mDice reported as mean{\scriptsize$\pm$std}). Classes: G~=~Granulation, S~=~Slough/Fibrin, N~=~Necrosis. Best result highlighted.
\textit{Params} denotes the number of model parameters in millions.
DLV3+ stands for DeepLabV3+, DLV3 stands for DeepLabV3.
}
\label{tab:full}
\centering
\footnotesize
\renewcommand{\arraystretch}{1.05}
\setlength{\tabcolsep}{4pt}
\resizebox{\linewidth}{!}{%
\begin{tabular}{@{}llccccrrr@{}}
\toprule
\textbf{Architecture} & \textbf{Backbone} & \textbf{Weights}
  & \textbf{Params} & \textbf{mIoU} & \textbf{mDice}
  & \textbf{IoU\textsubscript{G}} & \textbf{IoU\textsubscript{S}}
  & \textbf{IoU\textsubscript{N}} \\
\midrule
\rowcolor{best}
MA-Net          & MiT-B2           & ImageNet &  34.97 & \textbf{0.7263}{\scriptsize$\pm$.0116} & \textbf{0.8407}{\scriptsize$\pm$.0075} & 0.7789 & 0.7042 & 0.6959 \\
U-Net           & MiT-B1           & ImageNet &  16.43 & 0.7203{\scriptsize$\pm$.0042} & 0.8367{\scriptsize$\pm$.0028} & 0.7778 & 0.7001 & 0.6831 \\
U-Net           & MiT-B2           & ImageNet &  27.48 & 0.7176{\scriptsize$\pm$.0166} & 0.8348{\scriptsize$\pm$.0111} & 0.7739 & 0.6976 & 0.6814 \\
U-Net           & ResNet-50        & SSL      &  32.52 & 0.7153{\scriptsize$\pm$.0145} & 0.8330{\scriptsize$\pm$.0097} & 0.7736 & 0.7060 & 0.6662 \\
MA-Net          & ResNet-34        & ImageNet &  31.78 & 0.7153{\scriptsize$\pm$.0207} & 0.8331{\scriptsize$\pm$.0138} & 0.7719 & 0.6969 & 0.6769 \\
DLV3       & ResNet-50        & SWSL     &  39.63 & 0.7148{\scriptsize$\pm$.0151} & 0.8328{\scriptsize$\pm$.0100} & 0.7707 & 0.6970 & 0.6766 \\
U-Net           & ResNet-34        & ImageNet &  24.44 & 0.7135{\scriptsize$\pm$.0202} & 0.8319{\scriptsize$\pm$.0136} & 0.7716 & 0.6959 & 0.6732 \\
DLV3+      & MiT-B3           & ImageNet &  45.23 & 0.7131{\scriptsize$\pm$.0152} & 0.8317{\scriptsize$\pm$.0102} & 0.7691 & 0.6897 & 0.6805 \\
DLV3+      & ResNet-50        & SWSL     &  26.68 & 0.7119{\scriptsize$\pm$.0087} & 0.8309{\scriptsize$\pm$.0056} & 0.7671 & 0.6910 & 0.6775 \\
UNet++          & ResNet-50        & ImageNet &  48.99 & 0.7115{\scriptsize$\pm$.0143} & 0.8306{\scriptsize$\pm$.0096} & 0.7696 & 0.6987 & 0.6663 \\
LinkNet         & ResNet-34        & ImageNet &  21.77 & 0.7114{\scriptsize$\pm$.0173} & 0.8305{\scriptsize$\pm$.0115} & 0.7685 & 0.6944 & 0.6712 \\
U-Net           & DenseNet-121     & ImageNet &  13.61 & 0.7110{\scriptsize$\pm$.0172} & 0.8302{\scriptsize$\pm$.0117} & 0.7647 & 0.6986 & 0.6697 \\
FPN             & MiT-B2           & ImageNet &  26.08 & 0.7108{\scriptsize$\pm$.0145} & 0.8301{\scriptsize$\pm$.0099} & 0.7677 & 0.6909 & 0.6739 \\
DLV3+      & ResNet-50        & ImageNet &  26.68 & 0.7100{\scriptsize$\pm$.0120} & 0.8296{\scriptsize$\pm$.0080} & 0.7631 & 0.6937 & 0.6731 \\
U-Net           & ResNet-18        & ImageNet &  14.33 & 0.7099{\scriptsize$\pm$.0137} & 0.8294{\scriptsize$\pm$.0090} & 0.7673 & 0.6958 & 0.6666 \\
UNet++          & DenseNet-121     & ImageNet &  30.07 & 0.7096{\scriptsize$\pm$.0116} & 0.8292{\scriptsize$\pm$.0078} & 0.7693 & 0.7008 & 0.6588 \\
FPN             & ResNeXt-50 & ImageNet &  25.59 & 0.7095{\scriptsize$\pm$.0162} & 0.8292{\scriptsize$\pm$.0109} & 0.7655 & 0.6958 & 0.6672 \\
U-Net           & ResNet-50        & SWSL     &  32.52 & 0.7084{\scriptsize$\pm$.0114} & 0.8282{\scriptsize$\pm$.0077} & 0.7711 & 0.6972 & 0.6569 \\
U-Net           & ResNet-50        & ImageNet &  32.52 & 0.7083{\scriptsize$\pm$.0115} & 0.8284{\scriptsize$\pm$.0078} & 0.7641 & 0.6936 & 0.6670 \\
FPN             & ResNet-34        & ImageNet &  23.16 & 0.7068{\scriptsize$\pm$.0223} & 0.8271{\scriptsize$\pm$.0151} & 0.7682 & 0.6844 & 0.6678 \\
UNet++          & ResNet-34        & ImageNet &  26.08 & 0.7068{\scriptsize$\pm$.0269} & 0.8272{\scriptsize$\pm$.0183} & 0.7613 & 0.6861 & 0.6729 \\
PAN             & ResNet-34        & ImageNet &  21.48 & 0.7053{\scriptsize$\pm$.0135} & 0.8263{\scriptsize$\pm$.0093} & 0.7618 & 0.6858 & 0.6682 \\
U-Net           & EfficientNet-B4  & ImageNet &  20.23 & 0.7046{\scriptsize$\pm$.0179} & 0.8257{\scriptsize$\pm$.0122} & 0.7634 & 0.6901 & 0.6602 \\
FPN             & ResNet-50        & ImageNet &  26.12 & 0.7043{\scriptsize$\pm$.0089} & 0.8257{\scriptsize$\pm$.0059} & 0.7612 & 0.6906 & 0.6612 \\
U-Net           & EfficientNet-B3  & ImageNet &  13.16 & 0.7036{\scriptsize$\pm$.0100} & 0.8249{\scriptsize$\pm$.0070} & 0.7681 & 0.6963 & 0.6465 \\
MA-Net          & ResNet-50        & ImageNet & 147.44 & 0.7030{\scriptsize$\pm$.0105} & 0.8245{\scriptsize$\pm$.0069} & 0.7661 & 0.6929 & 0.6499 \\
FPN             & EfficientNet-B4  & ImageNet &  19.35 & 0.7016{\scriptsize$\pm$.0115} & 0.8236{\scriptsize$\pm$.0080} & 0.7654 & 0.6888 & 0.6506 \\
LinkNet         & ResNet-50        & ImageNet &  31.18 & 0.7015{\scriptsize$\pm$.0092} & 0.8237{\scriptsize$\pm$.0061} & 0.7609 & 0.6878 & 0.6558 \\
U-Net           & MobileNetV2      & ImageNet &   6.63 & 0.7014{\scriptsize$\pm$.0097} & 0.8236{\scriptsize$\pm$.0064} & 0.7613 & 0.6897 & 0.6534 \\
U-Net           & EfficientNet-B0  & ImageNet &   6.25 & 0.7012{\scriptsize$\pm$.0067} & 0.8234{\scriptsize$\pm$.0044} & 0.7622 & 0.6860 & 0.6555 \\
DLV3       & ResNet-50        & ImageNet &  39.63 & 0.7012{\scriptsize$\pm$.0108} & 0.8234{\scriptsize$\pm$.0070} & 0.7614 & 0.6860 & 0.6562 \\
PSPNet          & ResNet-50        & ImageNet &  24.31 & 0.6960{\scriptsize$\pm$.0146} & 0.8198{\scriptsize$\pm$.0099} & 0.7531 & 0.6814 & 0.6534 \\
PSPNet          & ResNet-34        & ImageNet &  21.45 & 0.6934{\scriptsize$\pm$.0247} & 0.8178{\scriptsize$\pm$.0170} & 0.7537 & 0.6720 & 0.6546 \\
UNet++          & EfficientNet-B3  & ImageNet &  13.63 & 0.6924{\scriptsize$\pm$.0081} & 0.8168{\scriptsize$\pm$.0052} & 0.7645 & 0.6866 & 0.6260 \\
MA-Net          & EfficientNet-B3  & ImageNet &  17.17 & 0.6883{\scriptsize$\pm$.0042} & 0.8137{\scriptsize$\pm$.0033} & 0.7647 & 0.6807 & 0.6194 \\
DLV3+      & EfficientNet-B3  & ImageNet &  11.68 & 0.6862{\scriptsize$\pm$.0078} & 0.8127{\scriptsize$\pm$.0054} & 0.7517 & 0.6763 & 0.6306 \\
PAN             & EfficientNet-B3  & ImageNet &  10.81 & 0.6772{\scriptsize$\pm$.0094} & 0.8060{\scriptsize$\pm$.0063} & 0.7495 & 0.6683 & 0.6138 \\
\bottomrule
\end{tabular}}
\end{table}

\begin{figure}[tb]
\centering
\includegraphics[width=0.99\linewidth]{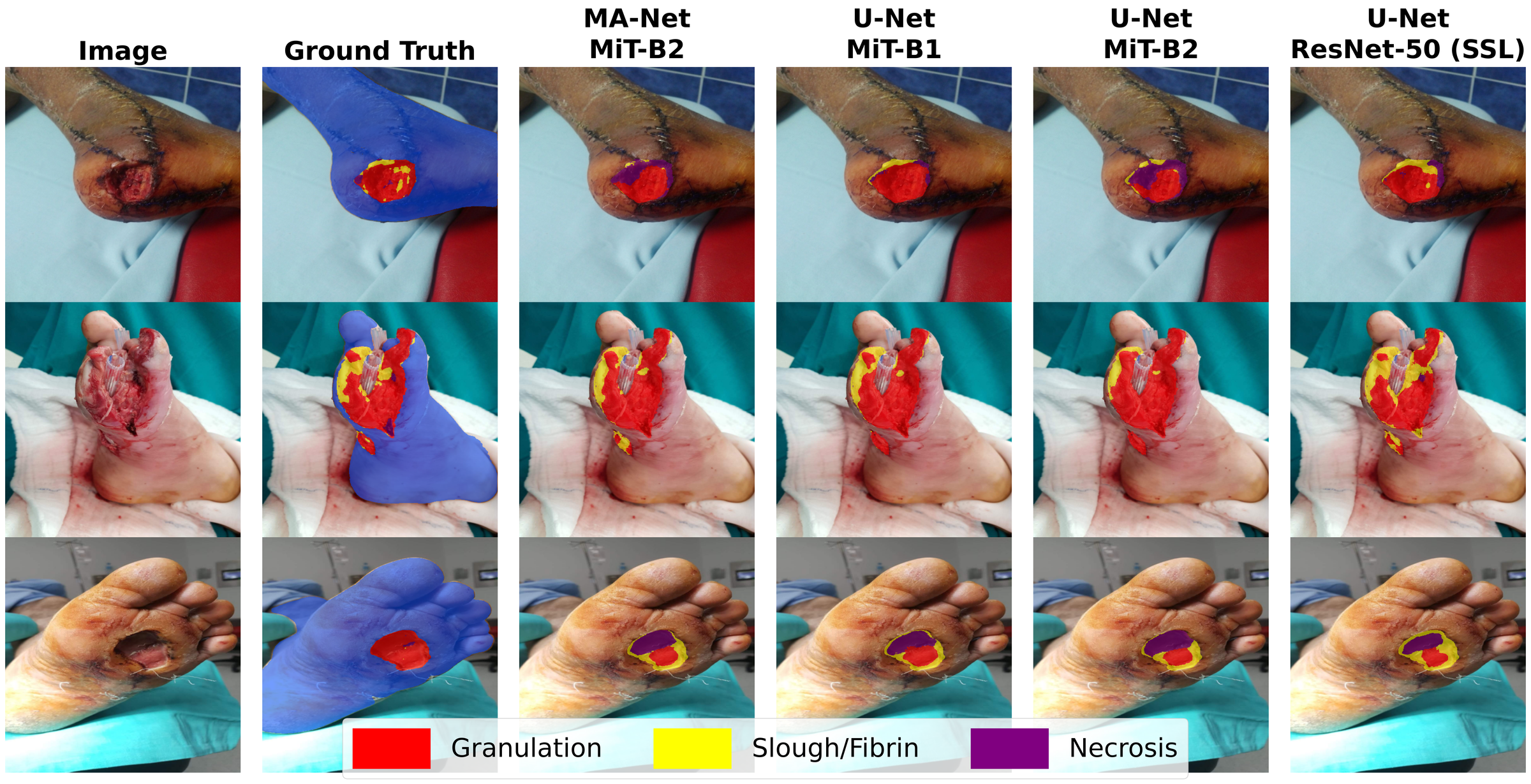}
\caption{Qualitative comparison of intra-wound tissue segmentation on three samples from the clinical dataset.
Each row shows the original wound image, the ground-truth annotation, and Stage~2 predictions from four top-performing configurations:
MA-Net \& MiT-B2, U-Net \& MiT-B1, U-Net \& MiT-B2, and U-Net \& ResNet-50 (SSL).
Colour coding: red~=~granulation, yellow~=~slough/fibrin, purple~=~necrosis, blue~=~skin remnant (void).}
\label{fig:qual_comparison}
\end{figure}

A qualitative comparison of tissue segmentation outputs across representative
model configurations is shown in \cref{fig:qual_comparison}.

\subsection{Impact of Encoder Backbone}

Grouping the 37 configurations from \cref{tab:full} by encoder backbone family
reveals a clear performance hierarchy.
Mix Transformer (MiT) encoders (5~configurations) achieve the highest mean
mIoU of 0.7176, with the best individual result of 0.7263 (MA-Net\&MiT-B2). The MiT advantage suggests that the hierarchical self-attention mechanism, which
captures long-range tissue boundary dependencies that CNN backbones miss. 
DenseNet follows with a mean IoU of 0.7103 across 2 configurations (best: 0.7110),
and 
ResNet encoders (20 configurations) reach a mean IoU of 0.7074 with a best of 0.7153,
demonstrating varying performance across the widest range of architectures.
EfficientNet encoders
(8 configurations) rank lowest with a mean of 0.6944 due to their lower number of parameters on average.

Results in \cref{tab:wound_loc_benchmark} (wound region segmentation) reveal a similar observation where MiT encoders outperform ResNet and others on average.

\subsection{Impact of Segmentation Architecture}

Grouping configurations from \cref{tab:full} by network architecture reveals that
architecture choice has a modest effect compared to encoder backbone selection.
Across all 9 architectures, the spread of mean IoU is narrow (0.6913 to 0.7096):
U-Net leads with a mean IoU of 0.7096 across 12 configurations (best: 0.7203);
MA-Net ranks second at 0.7082 mean (best: 0.7263, the overall top result).


A similar pattern holds for wound region segmentation (Tab.~\ref{tab:wound_loc_benchmark}): across the five architectures evaluated, the spread of mean IoU is even narrower, ranging from 0.8868 (FPN) to 0.8968 (U-Net++), further confirming that architecture choice contributes little relative to encoder backbone selection.In summary, for both tasks, the choice of encoder backbone dominates over the choice of segmentation architecture.

\subsection{Effect of Pretrained Weights}
The semi-supervised pretrained weights (SSL) consistently outperform standard ImageNet pretraining (especially observe U-Net\&ResNet-50 pairs in Tables \ref{tab:wound_loc_benchmark} and \ref{tab:full}). The gains are aroun 0.7 percentage points at mIoU. 

Combined supervised and semi-supervised learning weights (SWSL) also show improvement over ImageNet pretraining for the U-Net\&ResNet-50 pairs, but the improvement is marginal (0.1\% or less) compared to the improvement gained by SSL. Peculiar to the wound region segmentation task (Table \ref{tab:wound_loc_benchmark}), SWSL brings a more significant improvement for DeepLabV3+ \& ResNet-50 pair alone.

These improvements remain modest compared to the gap driven by encoder architecture: CNN-based configurations even with the best available pretrained weights still trail the best MiT-based configuration, reinforcing that encoder choice outweighs pretraining strategy for this task.

\subsection{Per-Class Performance Analysis}

Per-class results of intra-wound segmentation (cf. \cref{tab:full}), for the top-5 configurations, 
Granulation tissue is consistently the easiest to delineate across all models, achieving IoU above 0.77 for MiT-encoder models, attributable to its distinctive and spectrally consistent red appearance.
Slough/fibrin is moderately challenging; its yellowish coloration can overlap visually with skin at the wound margin.
Necrosis exhibits the greatest variability and is universally the hardest class: its appearance ranges from dark brown to black and varies with
wound depth, desiccation level, and illumination angle.
Notably, MA-Net~\&~MiT-B2 achieves the highest individual necrosis IoU among the top configurations (0.6959) reinforcing that the strongest overall encoder also handles the most texturally complex class best.

\subsection{Discussion}

MiT encoders outperform CNN backbones on average across all tested segmentation architectures.
We attribute this to their global self-attention mechanism, which captures
clinically context-dependent tissue boundaries from the first stage, and to
their hierarchical multi-scale feature maps that align with the multi-scale structure of wound tissue.
SSL pretrained weights provide modest gains over ImageNet pretrained weights, but are insufficient to close the gap to MiT encoders, suggesting that architectural choice outweighs pretrained weight strategy for this task.

\section{Conclusions}
\label{sec:conclusions}

This paper introduced a clinical wound tissue dataset and a systematic benchmark spanning multiple segmentation architectures, encoder backbones, and pretrained weight strategies. The central finding is that encoder backbone choice is the dominant performance driver: Vision Transformer backbones (MiT series) outperform all CNN families regardless of segmentation architecture, with the MA-Net \& MiT-B2 combination achieving the best overall result. Pretrained weight strategy plays a secondary role, with semi-supervised weight variants offering modest but consistent gains over standard ImageNet initialisation for CNN encoders. For practitioners, a lighter MiT-based U-Net configuration offers the best accuracy–efficiency trade-off, while lightweight CNN encoders such as EfficientNet-B0 suit resource-constrained deployment settings. Going forward, we plan to expand the dataset with additional clinical sites and introduce new tissue classes such as tendon and bone exposure to better reflect the full spectrum of wound severity.

%
%
\bibliographystyle{splncs04}
\bibliography{references}
\end{document}